\pdfoutput=1

\documentclass[11pt]{article}

\usepackage{emnlp2021}

\usepackage{times}
\usepackage{latexsym}

\usepackage[T1]{fontenc}

\usepackage[utf8]{inputenc}

\usepackage{microtype}

\title{AligNART: Non-autoregressive Neural Machine Translation by Jointly Learning to Estimate Alignment and Translate}

\author{Jongyoon Song$^{1, 2}$\Thanks{\hspace{0.1em} This work was done during an internship at Kakao Enterprise.} \And Sungwon Kim$^{1}$ \\
    $^{1}$Data Science and AI Laboratory, Seoul National University, South Korea \\
    $^{2}$Kakao Enterprise, South Korea \\
    $^{3}$Interdisciplinary Program in Artificial Intelligence, Seoul National University, South Korea \\
    \texttt{\{coms1580, ksw0306, sryoon\}@snu.ac.kr} \\
    \And Sungroh Yoon$^{1, 3}$\Thanks{\hspace{0.1em} Corresponding author.} \\
}

\begin{document}
\maketitle
\begin{abstract}
Non-autoregressive neural machine translation (NART) models suffer from the \textit{multi-modality problem} which causes translation inconsistency such as token repetition.
Most recent approaches have attempted to solve this problem by implicitly modeling dependencies between outputs.
In this paper, we introduce AligNART, which leverages full alignment information to explicitly reduce the modality of the target distribution.
AligNART divides the machine translation task into $(i)$ alignment estimation and $(ii)$ translation with aligned decoder inputs, guiding the decoder to focus on simplified one-to-one translation.
To alleviate the alignment estimation problem, we further propose a novel \textit{alignment decomposition} method.
Our experiments show that AligNART outperforms previous non-iterative NART models that focus on explicit modality reduction on WMT14 En$\leftrightarrow$De and WMT16 Ro$\rightarrow$En.
Furthermore, AligNART achieves BLEU scores comparable to those of the state-of-the-art connectionist temporal classification based models on WMT14 En$\leftrightarrow$De.
We also observe that AligNART effectively addresses the token repetition problem even without sequence-level knowledge distillation.
\end{abstract}

\section{Introduction}\label{sec:introduction}
In the neural machine translation (NMT) domain, non-autoregressive NMT (NART) models (\citealp{gu-2018-nonautoregressive}) have been proposed to alleviate the low translation speeds of autoregressive NMT (ART) models.
However, these models suffer from degenerated translation quality (\citealp{gu-2018-nonautoregressive}; \citealp{zhiqing2019CRF}).
To improve the translation quality of NART, several studies on NART iteratively refine decoded outputs with minimal iterations (\citealp{ghazvininejad-etal-2019-mask}; \citealp{kasai2020non}; \citealp{lee-etal-2020-iterative}; \citealp{guo-etal-2020-jointly}; \citealp{saharia-etal-2020-non}); other recent works target to improve NART without iteration (\citealp{qian-etal-2021-glancing}; \citealp{gu-kong-2021-fully}). 

One of the significant limitations of non-iterative NART models is the \textit{multi-modality problem}. 
This problem originates from the fact that the models should maximize the probabilities of multiple targets without considering conditional dependencies between target tokens.
For example, in English-to-German translation, a source sentence "Thank you very much." can be translated to "Danke sch\"on." or "Vielen Dank.".
Under the conditional independence assumption, the non-iterative NART models are likely to generate improper translations such as "Danke Dank." or "Vielen sch\"on." (\citealp{gu-2018-nonautoregressive}). 
For the same reason, other inconsistency problems such as token repetition or omission occur frequently in non-iterative NART (\citealp{gu-kong-2021-fully}).

There are two main methods for non-iterative NART to address the \textit{multi-modality problem}.
Some works focus on an implicit modeling of the dependencies between the target tokens (\citealp{gu-kong-2021-fully}).
For example, \citet{ghazvininejad2020aligned}, \citet{saharia-etal-2020-non}, and \citet{gu-kong-2021-fully} modify the objective function based on dynamic programming, whereas \citet{qian-etal-2021-glancing} provide target tokens to the decoder during training.

On the other hand, other works focus on an explicit reduction of the modality of the target distribution by utilizing external source or target sentence information rather than modifying the objective function.
For example, \citet{akoury-etal-2019-syntactically} and \citet{liu-etal-2021-enriching} use syntactic or semantic information; \citet{gu-2018-nonautoregressive}, \citet{zhou2020non}, and \citet{Ran_2021_guiding} use the alignment information between source and target tokens.
However, previous explicit modality reduction methods show suboptimal performance.

\citet{zhou2020non} and \citet{Ran_2021_guiding} extract fertility (\citealp{brown1993mathematics}) and ordering information in word alignments, which enables the modeling of several types of mappings except for many-to-one and many-to-many cases.
We hypothesize that leveraging entire mappings significantly reduces the modality and is the key to performance improvement.

In this work, we propose AligNART, a non-iterative NART model that mitigates the \textit{multi-modality problem} by utilizing complete information in word alignments.
AligNART divides the machine translation task into $(i)$ alignment estimation and $(ii)$ non-autoregressive translation under the given alignments.
Modeling all the type of mapping guides $(ii)$ more close to one-to-one translation.
In AligNART, a module called Aligner is simply augmented to NAT (\citealp{gu-2018-nonautoregressive}) which estimates alignments to generate aligned decoder inputs.

However, it is challenging to estimate the complex alignment information using only source sentence during inference.
Specifically, Aligner should simultaneously predict the number of target tokens corresponding to each source token and their mapping.
To overcome this problem, we further propose \textit{alignment decomposition} which factorizes the alignment process into three sub-processes: \textit{duplication}, \textit{permutation}, and \textit{grouping}.
Each sub-process corresponds to much feasible sub-problems: one-to-many mapping, ordering, and many-to-one mapping, respectively.

Our experimental results show that AligNART outperforms previous non-iterative NART models of explicit modality reduction on WMT14 En$\leftrightarrow$De and WMT16 Ro$\rightarrow$En.
AligNART achieves performance comparable to that of the recent state-of-the-art non-iterative NART model on WMT14 En$\leftrightarrow$De.
We observe that the modality reduction in AligNART addresses the token repetition issue even without sequence-level knowledge distillation (\citealp{kim-rush-2016-sequence}).
We also conduct quantitative and qualitative analyses on the effectiveness of \textit{alignment decomposition}.
\section{Background}\label{sec:background}
Given a source sentence $x = \{x_1, x_2, ..., x_M\}$ and its translation $y = \{y_1, y_2, ..., y_N\}$, ART models with encoder-decoder architecture are trained with chained target distributions and infer the target sentence autoregressively:
\begin{equation}
    p(y|x) = \prod_{n=1}^{N} p(y_n|y_{<n},x).
\end{equation}
At each decoding position $n$, the decoder of the model is conditioned with previous target tokens $y_{<n} = \{y_1, ..., y_{n-1}\}$, which is the key factor of performance in ART models. 
Previous target tokens reduce the target distribution modality and provide information about the target sentence. 
However, the autoregressive decoding scheme enforces the decoder to iterate $N$ times to complete the translation and increases the translation time linearly with respect to the length of the target sentence.

Non-iterative NART models (\citealp{gu-2018-nonautoregressive}; \citealp{zhiqing2019CRF}; \citealp{sun2020approach}) assume conditional independence between the target tokens to improve the translation speed:
\begin{equation}\label{eq:NAT_objective}
    p(y|x) = p(N|x) \cdot \prod_{n=1}^{N} p(y_n|x),
\end{equation}
where $N$ is the predicted target length to parallelize the decoding process. 
Non-iterative NART models provide only the length information of the target sentence to the decoder, which is insufficient to address the \textit{multi-modality problem}.
\begin{figure*}[t!]
\sbox\twosubbox{%
  \resizebox{\dimexpr0.9\textwidth-1em}{!}{%
    \includegraphics[height=3cm]{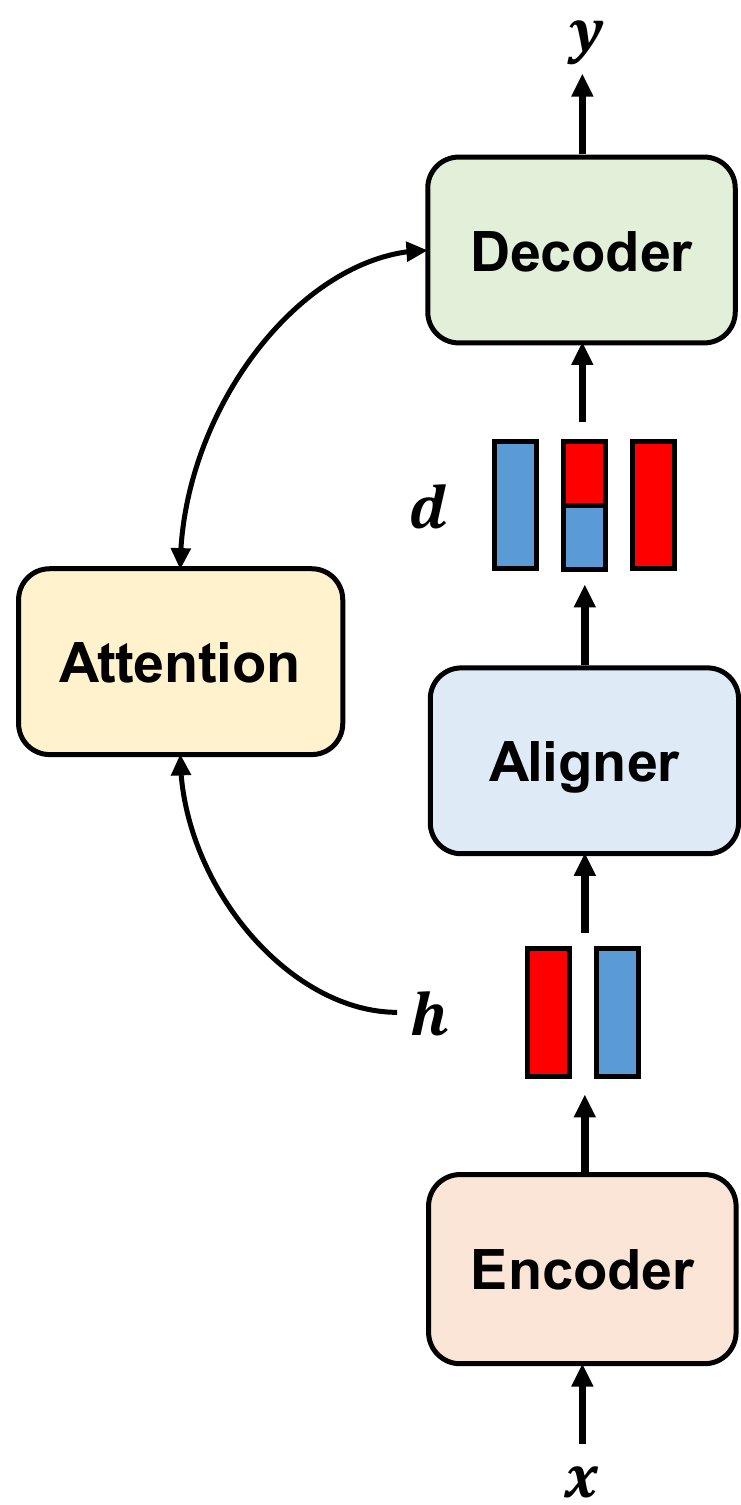}%
    \includegraphics[height=3cm]{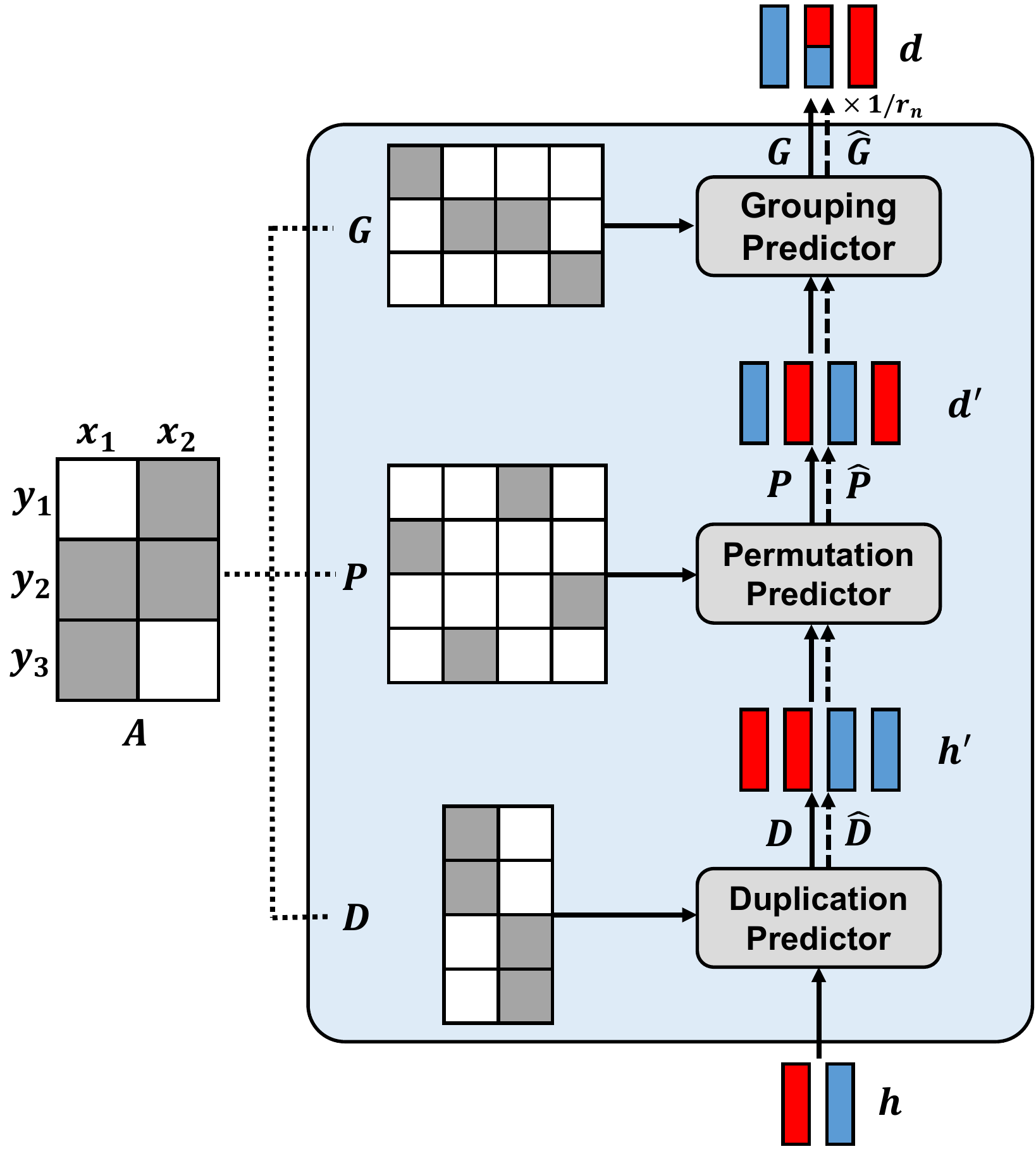}%
  }%
}
\setlength{\twosubht}{\ht\twosubbox}

\centering
\subcaptionbox{Model overview of AligNART\label{fig:1a}}{%
  \includegraphics[height=0.85\twosubht]{figures/Figure1a.pdf}%
}\hspace{15mm}
\subcaptionbox{Alignment decomposition and sub-processes of Aligner\label{fig:1b}}{%
  \includegraphics[height=0.85\twosubht]{figures/Figure1b.pdf}%
}
\caption{(a) Given the encoder outputs $h$, Aligner returns aligned encoder outputs $d$. The decoder then translates the aligned inputs to $y$. (b) The dotted lines indicate the \textit{alignment decomposition}. During training, predictors are trained with the decomposed matrices D, P, and G, and align inputs using the ground truth as indicated by the solid lines. During inference, predictors align inputs using the estimated matrices as indicated by the dashed lines.}\label{fig:overall_architecture_of_alignart}
\end{figure*}

\section{AligNART}\label{sec:AligNART}
\subsection{Model Overview}\label{subsec:model_overview}
Given the word alignments between the source and target sentences $A \in {\{0,1\}}^{N \times M}$, we factorize the task into $(i)$ alignment estimation and $(ii)$ translation with aligned decoder inputs as follows:
\begin{equation}\label{eq:aligner}
    p(y|x) = p(A|x) \cdot \prod_{n=1}^{N} p(y_n|x, A),
\end{equation}
where $M$ and $N$ are the lengths of the source and target sentences, respectively.
Although we can also modify the negative log-likelihood loss to model dependencies between outputs such as connectionist temporal classification (CTC) loss (\citealp{graves-2006-connectionist}), we focus on the effect of the introduction of alignment as additional information.
AligNART is based on the encoder-decoder architecture, with an alignment estimation module called Aligner as depicted in Figure \ref{fig:1a}.
The encoder maps the embedding of the source tokens into hidden representations $h=\{h_1, h_2, ..., h_M\}$.
Aligner constructs the aligned decoder inputs $d=\{d_1, d_2, ..., d_N\}$ as follows: 
\begin{equation}\label{eq:aligned_decoder_input}
    d_n = \frac{1}{r_n}\sum_{m=1}^{M}A_{n,m}\cdot h_m.
\end{equation}
where $r_n$ is the number of non-zero elements in the $n$-th row of $A$.
Given the aligned decoder inputs, the decoder is guided to focus on a one-to-one translation from $d_n$ to $y_n$.
One-to-one mapping significantly reduces the modality of the target distribution.

The key component of AligNART, Aligner, models a conditional distribution of alignments $A$ given the source sentence $x$ during training, and aligns encoder outputs using the estimated alignments during inference, as depicted in Figure \ref{fig:1b}.
The ground truth of the alignments is extracted using an external word alignment tool.
However, alignment estimation given only the source sentence is challenging since the alignment consists of two components related with target tokens:
\begin{itemize}
    \item The number of target tokens that correspond to each encoder output $h_m$.
    \item The positions of the target tokens to which $h_m$ corresponds.
\end{itemize}
The Aligner decomposes the alignment for effective estimation, which is described in Section \ref{subsec:aligner}.

\subsection{Aligner}\label{subsec:aligner}
To alleviate the alignment estimation problem, we start by factorizing the alignment process as shown in Figure \ref{fig:1b}. 
First, we copy each encoder output $h_m$ by the number of target tokens mapped to $h_m$, which is denoted as $c_m=\sum_{n}{A_{n,m}}$. 
Given the duplicated encoder outputs $h'$, we have to predict the positions of target tokens to which each element in $h'$ is mapped.

We further decompose the remaining prediction process into \textit{permutation} and \textit{grouping}, since non-iterative NART models have no information about the target length $N$ during inference.
In the permutation process, $h'$ is re-ordered into $d'$ such that elements corresponding to the same target token are placed adjacent to each other.
In the grouping process, each element in $d'$ is clustered into $N$ groups by predicting \textit{whether each element is mapped to the same target token as the previous element}.
$r_n=\sum_{m}{A_{n,m}}$ denotes the number of elements in the $n$-th group which is equivalent to $r_n$ in Equation \ref{eq:aligned_decoder_input}.
Finally, we can derive the decoder inputs $d$ in Equation \ref{eq:aligned_decoder_input} by averaging the elements in each group in $d'$.
In summary, we decompose the alignment estimation task into three sequential sub-tasks: \textit{duplication}, \textit{permutation}, and \textit{grouping}.

\subsubsection{Alignment Decomposition}\label{subsubsec:definition_and_validation_of_decomposition} 
As shown in Figure \ref{fig:1b}, we factorize the alignment matrix $A$ into duplication, permutation, and grouping matrices that correspond to each process.
$h'=\{h_{1,1}, ...,h_{1,c_1},...,h_{M,1},...,d_{M,c_M}\}$ denotes the duplicated encoder outputs where $h_{i,j}$ is the $j$-th copied element of $h_i$.
Similarly, $d'=\{d_{1,1}, ...,d_{1,r_1},...,d_{N,1},...,d_{N,r_N}\}$ denotes the permuted encoder outputs where $d_{i,j}$ is the $j$-th element in the $i$-th group.
The number of non-zero elements in the alignment matrix is defined as $L=\sum_{m}{c_m}=\sum_{n}{r_n}$.

\textbf{Duplication Matrix} Aligner copies $h_m$ by $c_m$ to construct the duplicated encoder outputs $h'$ with a duplication matrix $D\in {\{0,1\}}^{L\times M}$.
Let $C_m=\sum_{i=1}^{m}c_i$ and $C_0=0$. Then, we can define $D$ using $c_m$ as follows:
\begin{equation}\label{eq:duplication_matrix}
    D_{l,m} = 
    \begin{cases}
        1 & \text{if } C_{m-1} < l \le C_m \\
        0 & \text{else}.
    \end{cases}
\end{equation}
We index $h'$ by the following rule:
\begin{itemize}
\item For any $h_{m,i}$ and $h_{m,j}$ ($i<j$), which are matched to $d_{x_i,y_i}$ and $d_{x_j,y_j}$, respectively, $x_i \le x_j$ and $y_i \le y_j$.
\end{itemize}
The duplication matrix $D$ contains similar information to fertility (\citealp{gu-2018-nonautoregressive}).

\textbf{Permutation Matrix} Aligner re-orders $h'$ to construct $d'$ with a permutation matrix $P\in {\{0,1\}}^{L\times L}$.
Since all the indexed elements in $h'$ and $d'$ are distinct, the permutation matrix $P$ is uniquely defined.

\textbf{Grouping Matrix} Aligner finally aggregates $d'$ to construct $d$, the aligned decoder inputs, with a grouping matrix $G\in {\{0,1\}}^{N\times L}$.
Let $R_n=\sum_{i=1}^{n}r_i$ and $R_0=0$. Then, $G$ can be defined using $r_n$ as follows:
\begin{equation}\label{eq:grouping_matrix}
    G_{n,l} = 
    \begin{cases}
        1 & \text{if } R_{n-1} < l \le R_n \\
        0 & \text{else}.
    \end{cases}
\end{equation}
We index $d'$ by the following rule:
\begin{itemize}
\item For any $d_{n,i}$ and $d_{n,j}$ ($i<j$), which are matched to $h_{x_i,y_i}$ and $h_{x_j,y_j}$, respectively, $x_i \le x_j$ and $y_i \le y_j$.
\end{itemize}
We can derive the aligned decoder inputs by separately estimating the decomposed matrices $D$, $P$, and $G$, which approximately correspond to one-to-many mapping, ordering, and many-to-one mapping, respectively.
The decomposed matrices have an easily predictable form while recovering the complete alignment matrix.

\subsubsection{Training}\label{subsubsec:training}
Aligner consists of three prediction sub-modules: duplication, permutation, and grouping predictors.
Each of them estimates the decomposed alignment matrix as follows:
\begin{equation}
    p(A|x) = p(G|x, P, D)\cdot p(P|x, D)\cdot p(D|x).
\end{equation}
The duplication predictor learns to classify the number of copies of $h_m$.
The duplication loss is defined as follows:
\begin{equation}\label{eq:duplication_loss}
    \mathcal{L}_D = -{\frac{1}{M}}\sum_{m=1}^{M}\log p_m(c_m),
\end{equation}
where $p_m$ is the predicted probability distribution of the duplication at the position $m$.
To discriminate copied elements in $h'$, we add copy position embedding to $\{h_{m,1}, ..., h_{m,c_m}\}$ for the next two predictors.

The permutation predictor takes the duplicated encoder outputs $h'$ as inputs.
We simplify the permutation prediction problem into a classification of the re-ordered position.
For the permutation loss, we minimize the KL-divergence between the prediction $P^{pred}$ and the ground truth $P^{GT}$.
\begin{equation}\label{eq:permutation_loss}
    \mathcal{L}_P = -{\frac{1}{L}}\sum_i\sum_j P_{i,j}^{GT}\log P_{i,j}^{pred}.
\end{equation}
Given the permuted encoder outputs, the grouping predictor conducts a binary classification task of whether $d'_l$ is assigned to the same group as $d'_{l-1}$.
Let the label at the position $l$ be $g_l$. Then, we define $g_l$ from $G$ as follows:
\begin{equation}\label{eq:G_and_g}
     g_{l} = 
    \begin{cases}
        1 & \text{if } G_{\ast,l} = G_{\ast,l-1} \text{ and } l > 1 \\
        0 & \text{else}.
    \end{cases}
\end{equation}
The grouping loss is defined as follows:
\begin{equation}\label{eq:grouping_loss}
    \mathcal{L}_G = -{\frac{1}{L}}\sum_{l=1}^{L}\log p_l(g_l),
\end{equation}
where $p_l$ is the predicted probability distribution of the grouping predictor at position $l$.

Our final loss function is defined as the sum of the negative log-likelihood based translation loss $\mathcal{L}_{T}$ and alignment loss $\mathcal{L}_{A}$:
\begin{equation}\label{eq:overall_loss}
    \mathcal{L} = \mathcal{L}_{T} + \mathcal{L}_{A} = \mathcal{L}_{T} + \alpha\mathcal{L}_D + \beta\mathcal{L}_P + \gamma\mathcal{L}_G,
\end{equation}
where we set $\alpha=\beta=\gamma=0.5$ for all the experiments.

\subsubsection{Inference}\label{subsubsec:inference}
During inference, Aligner sequentially predicts the duplication, permutation, and grouping matrices to compute the aligned decoder inputs $d$ as depicted in Figure \ref{fig:1b}.
The duplication predictor in Aligner infers $\hat{c}_m$ at each position $m$; then, we can directly construct a duplication matrix $\hat{D}$ using Equation \ref{eq:duplication_matrix}.
The permutation predictor predicts the distribution of the target position $P^{pred}$.
We obtain a permutation matrix $\hat{P}$ that minimizes the KL-divergence as follows:
\begin{equation}
    \hat{P}=\argmin_{P} (-\sum_i\sum_j P_{i,j}\log P^{pred}_{i,j}).
\end{equation}
We utilize the \textit{linear sum assignment problem solver} provided by \citet{scipy} to find $\hat{P}$.
The grouping predictor infers the binary predictions $\hat{g_l}$ from the permuted encoder outputs.
We construct a grouping matrix $\hat{G}$ using $\hat{g_l}$ and Equations \ref{eq:grouping_matrix} and \ref{eq:G_and_g}.
With a predicted alignment matrix $\hat{A}=\hat{G}\cdot \hat{P}\cdot \hat{D}$, Aligner constructs the decoder inputs using Equation \ref{eq:aligned_decoder_input}, and the decoder performs translation from the aligned inputs.

\subsubsection{Decoding Strategies}
For the re-scoring based decoding method, we select candidates of alignments using the predicted distributions in the duplication and grouping predictors.

We identify $m'$ positions in the outputs of the duplication predictor, where the probability of the predicted class is low. 
We then construct a $2^{m'}$-candidate pool where the predictions in part of the $m'$ positions are replaced with the second probable class.
Next, we identify the top-$a$ candidates with the highest joint probabilities. 
Similarly, we construct a $2^{l'}$-candidate pool and identify $b$ candidates in the grouping predictor for the $a$ candidates.
Finally, we rank $a\cdot b$ translations for the alignments candidates using a teacher ART model and select the best translation among them.

\subsection{Architecture of AligNART}\label{subsec:architecture_of_alignart}
We use the deep-shallow (12-1 for short) Transformer (\citealp{vaswani-2017-attention}) architecture (i.e., 12-layer encoder and 1-layer decoder) proposed by \citet{kasai2020deep} for two reasons. 
First, a deeper encoder assists Aligner to increase the estimation accuracy of the alignment matrix during inference. 
Second, the deep-shallow architecture improves the inference speed since the encoder layer has no cross-attention module compared to the decoder layer.
The architecture of the duplication, permutation, and grouping predictor is shown in the Appendix.

\subsection{Alignment Score Filtering}
Some alignment tools such as GIZA++ (\citealp{och-ney-2003-systematic}) provide an alignment score for each sentence pair as a default. 
Samples with low alignment scores are more likely to contain noise caused by sentence pairs or alignment tools.
For GIZA++, we filter out a fixed portion of samples with low alignment scores to ease the alignment estimation. 
Since the pair of long sentences tends to be aligned with a low score, we apply the same filtering portion for each target sentence length.

\begin{table*}
\centering
\resizebox{\linewidth}{!}{
\begin{tabular}{l|cccc|cccc}
\hline
\toprule
\multicolumn{1}{c}{ } &
\multicolumn{4}{c}{WMT14 En-De} &
\multicolumn{4}{c}{WMT16 En-Ro} \\
Models & En$\rightarrow$ & De$\rightarrow$ & Time & Speedup & En$\rightarrow$ & Ro$\rightarrow$ & Time & Speedup \\
\midrule
\multicolumn{9}{c}{\textit{Autoregressive Models}} \\
\hline
Transformer (\citealp{vaswani-2017-attention}) & 27.3 & - & - & - & - & - & - & - \\
Transformer (ours) & 27.4 & 31.4 & 314 & $\times$1.0 & 34.1 & 33.9 & 307 & $\times$1.0 \\
\hline
\multicolumn{9}{c}{\textit{Non-iterative Non-autoregressive Models (implicit dependency modeling)}} \\
\hline
FlowSeq (\citealp{ma-etal-2019-flowseq}) & 21.5 & 26.2 & - & - & 29.3 & 30.4 & - & - \\
AXE (\citealp{ghazvininejad2020aligned}) & 23.5 & 27.9 & - & - & 30.8 & 31.5 & - & - \\
NAT-EM (\citealp{sun2020approach}) & 24.5 & 27.9 & 24 & $\times$16.4 & - & - & - & - \\
NARLVM (\citealp{lee-etal-2020-iterative}) & 25.7 & - & 19 & $\times$15.0 & - & 28.4 & 18 & $\times$34.0 \\
GLAT (\citealp{qian-etal-2021-glancing}) & 25.2 & 29.8 & - & $\times$15.3 & 31.2 & 32.0 & - & $\times$15.3 \\
Imputer (\citealp{saharia-etal-2020-non}) & 25.8 & 28.4 & - & $\times18.6$ & 32.3 & 31.7 & - & - \\
CTC (\citealp{gu-kong-2021-fully}) & \textbf{26.5} & \textbf{30.5} & - & $\times$16.8 & \textbf{33.4} & \textbf{34.1} & - & $\times$16.8 \\
\hline
\multicolumn{9}{c}{\textit{Non-iterative Non-autoregressive Models (explicit modality reduction)}} \\
\hline
NAT-FT (\citealp{gu-2018-nonautoregressive})& 17.7 & 21.5 & 39 & $\times$15.6 & 27.3 & 29.1 & 39 & $\times$15.6 \\
Distortion (\citealp{zhou2020non})& 22.7 & - & - & - & 29.1 & - & - & - \\
ReorderNAT (\citealp{Ran_2021_guiding})& 22.8 & 27.3 & - & $\times$16.1 & 29.3 & 29.5 & - & $\times$16.1 \\
SNAT (\citealp{liu-etal-2021-enriching})& 24.6 & 28.4 & 27 & $\times$22.6 & \textbf{32.9} & 32.2 & 27 & $\times$22.6 \\
\hline
AligNART (FA, ours) & 25.7 & 29.1 & 23 & $\times$13.6 & 31.7 & 32.2 & 22 & $\times$13.9 \\
AligNART (GZ, ours) & \textbf{26.4} & \textbf{30.4} & 24 & $\times$13.4 & 32.5 & \textbf{33.1} & 24 & $\times$13.0 \\
\hline
\end{tabular}}
\caption{
BLEU scores and inference speed of baselines and our model on four translation tasks. 
\textit{Time} is an average sentence-wise latency in milliseconds. 
\textit{Speedup} is a relative speedup ratio compared to
the Transformer-based ART model with beam width 5.
}\label{table:overall_BLEU_scores}
\end{table*}

\section{Experimental Setups}\label{sec:experiments}
\subsection{Datasets and Preprocessing}\label{subsec:datasets_and_preprocessing}
We evaluate our method on two translation datasets: WMT14 English-German (En-De) and WMT16 English-Romanian (En-Ro). 
WMT14 En-De/WMT16 En-Ro datasets contain 4.5M/610K training pairs, respectively.

For WMT14 En-De dataset, we use preprocessing pipelines provided by \textit{fairseq}\footnote[1]{\url{https://github.com/pytorch/fairseq}} (\citealp{ott-etal-2019-fairseq}). 
For WMT16 En-Ro dataset, we use the preprocessed corpus provided by \citet{lee-etal-2018-deterministic}.
Preprocessed datasets share a vocabulary dictionary between the source and target languages.
We use \textit{fast align} (FA) (\citealp{dyer-etal-2013-a}) and GIZA++ (GZ), which is known to be more accurate than \textit{fast align}, as word alignment tools. 
All the corpus are passed to the alignment tools at the subword-level.
We filter out samples where the maximum number of duplications exceed 16.
We explain the details of the alignment processing in the Appendix.

We use the sequence-level knowledge distillation method (KD) for the distillation set. 
Transformer ART models are trained to generate the distillation set for each translation direction.

\subsection{Models and Baselines}\label{subsec:models_and_baselines}
We compare our model with several non-iterative NART baselines, and divide the non-iterative NART models into two types as aforementioned: \textit{implicit dependency modeling} and \textit{explicit modality reduction} (see Table \ref{table:overall_BLEU_scores}).
We also train the ART models and deep-shallow NAT for the analysis.
Our models are implemented based on \textit{fairseq}.

AligNART is implemented based on the deep-shallow Transformer architecture. 
We set $d_{model}/d_{hidden}$ to 512$/$2048 and the dropout rate to 0.3.
The number of heads in multi-head attention modules is 8 except for the last attention module of the permutation predictor which is 1. 
We set the batch size to approximately 64K tokens for all the models we implement.
All these models we implement are trained for 300K$/$50K steps on En-De$/$En-Ro datasets, respectively.
For AligNART, we average 5 checkpoints with the highest validation BLEU scores in the 20 latest checkpoints.

For optimization, we use Adam optimizer (\citealp{kingma2015adam}) with $\beta=(0.9, 0.98)$ and $\epsilon=10^{-8}$. The learning rate scheduling follows that of \citet{vaswani-2017-attention}, starting from $10^{-7}$ and warms up to $5e$-4 in 10K steps. 
We use the label smoothing technique with $\epsilon_{ls}=0.1$ for the target token distribution and each row of permutation matrix. 
The translation latency is measured on an NVIDIA Tesla V100 GPU.
\section{Results}\label{sec:results}

\begin{table}
\centering
\resizebox{\columnwidth}{!}{
\begin{tabular}{l|cc|cc}
\hline
\toprule
\multicolumn{1}{c}{ } &
\multicolumn{2}{c}{WMT14 En-De} &
\multicolumn{2}{c}{WMT16 En-Ro} \\
Models & En$\rightarrow$ & De$\rightarrow$ & En$\rightarrow$ & Ro$\rightarrow$ \\
\hline
FlowSeq (n=15) & 23.1 & 28.1 & 31.4 & 32.1 \\
NAT-EM (n=9) & 25.8 & 29.3 & - & - \\
GLAT (n=7) & 26.6 & \textbf{31.0} & 32.9 & 33.5 \\
\hline
ReorderNAT (n=7) & 24.7 & 29.1 & 31.2 & 31.4 \\
SNAT (n=9) & 26.9 & 30.1 & \textbf{34.9} & 33.1 \\
AligNART (FA, n=8) & 26.5 & 30.3 & 32.7 & 33.1 \\
AligNART (GZ, n=8) & \textbf{27.0} & \textbf{31.0} & 33.0 & \textbf{33.7} \\
\hline
\end{tabular}}
\caption{BLEU scores of non-iterative NART models with re-scoring decoding scheme of $n$ candidates.}
\label{table:re-scoring_results}
\end{table}

\begin{table*}
\centering
\resizebox{\linewidth}{!}{
\begin{tabular}{l|ccc|ccc}
\hline
\toprule
\multicolumn{1}{c}{ } &
\multicolumn{3}{c}{En$\rightarrow$De} &
\multicolumn{3}{c}{De$\rightarrow$En} \\
Models & D & P & G & D & P & G \\
\hline
AligNART (FA, w/o KD) & 0.76/0.85 & 0.55/0.74 & 0.96/0.98 & 0.77/0.84 & 0.59/0.74 & 0.96/0.98 \\
AligNART (FA, w/ KD) & 0.75/0.89 & 0.53/0.83 & 0.95/1.00 & 0.76/0.88 & 0.57/0.84 & 0.96/1.00 \\
\hline
AligNART (GZ, w/o KD) & 0.69/0.78 & 0.76/0.91 & 1.00/1.00 & 0.71/0.82 & 0.81/0.92 & 1.00/1.00 \\
AligNART (GZ, w/ KD) & 0.66/0.88 & 0.71/0.94 & 1.00/1.00 & 0.68/0.88 & 0.76/0.95 & 1.00/1.00 \\
\hline
\end{tabular}}
\caption{Duplication (D), permutation (P), and grouping (G) accuracy of Aligner on WMT14 En-De validation set. Accuracy on raw and distilled datasets are written on the left and right of slash, respectively.}
\label{table:accuracy_on_aligner_components}
\end{table*}

\begin{table*}
\centering
\resizebox{\linewidth}{!}{
\begin{tabular}{l|l|l}
\hline
\toprule
\multicolumn{2}{c|}{Source} & Denken Sie , dass die Medien zu viel vom PS\_ G erwarten ?\\
\multicolumn{2}{c|}{Reference} & Do you think the media expect too much of PS\_ G ? \\
\hline
\multicolumn{2}{c|}{NAT (12-1)} & Do you think that the expect expect much from the PS\_ G ?\\
\hline
\multirow{4}{*}{Ours} & Duplication & \hlcolor{whitered1}{Denken} \hlcolor{whitered2}{Denken} Sie , dass die Medien zu viel \hlcolor{whiteblue1}{vom} \hlcolor{whiteblue2}{vom} \hlcolor{whitegreen1}{PSG} \hlcolor{whitegreen2}{PSG} erwarten ?\\
& Permutation & Denken Sie \hlcolor{whitered2}{Denken} , dass die Medien \hlcolor{whitebrown}{erwarten} zu viel vom vom PSG PSG ?\\
& Grouping & Denken Sie Denken \hlcolor{whitegray}{, dass} die Medien erwarten zu viel vom vom PSG PSG ?\\
& Output & \hlcolor{whitered1}{Do} you \hlcolor{whitered2}{think} \hlcolor{whitegray}{that} the media \hlcolor{whitebrown}{expect} too much \hlcolor{whiteblue1}{from} \hlcolor{whiteblue2}{the} \hlcolor{whitegreen1}{PS\_} \hlcolor{whitegreen2}{G} ?\\
\hline
\end{tabular}}
\caption{Visualized translation example of deep-shallow NAT and AligNART (FA) on WMT14 De$\rightarrow$En validation set. "\_" stands for subword tokenization. Highlighted tokens in duplication, permutation, and grouping processes are modified by the each module of Aligner. Highlighted tokens in output correspond to the tokens highlighted with the same colors in the previous processes. Note that Aligner first applies mean pooling to convert subword-level encoder outputs into word-level, as explained in the Appendix.}
\label{table:qualitative_study}
\end{table*}

\begin{table}
\centering
\resizebox{\columnwidth}{!}{
\begin{tabular}{l|cc|cc}
\hline
\toprule
\multicolumn{1}{c}{ } &
\multicolumn{2}{c}{\textit{fast align}} &
\multicolumn{2}{c}{GIZA++} \\
Models & En$\rightarrow$ & De$\rightarrow$ & En$\rightarrow$ & De$\rightarrow$ \\
\hline
AligNART & 25.7 & 29.1 & 26.4 & 30.4 \\
- Infer with D=I & 15.5 & 18.1 & 11.5 & 15.2 \\
- Infer with P=I & 19.4 & 22.2 & 21.5 & 24.7 \\
- Infer with G=I & 21.9 & 27.1 & 26.4 & 30.4 \\
\hline
\end{tabular}}
\caption{BLEU scores of Aligner ablation study on WMT14 En-De test set.}
\label{table:alignment_estimation}
\end{table}

\subsection{Main Results}\label{subsec:main_results}
Table \ref{table:overall_BLEU_scores} shows the BLEU scores, translation latency and speedup on WMT14 En-De and WMT16 En-Ro. 
In explicit modality reduction, AligNART (FA) achieves higher BLEU scores than Distortion and ReorderNAT, which utilize the same alignment tool, since we leverage the entire alignment information rather than partial information such as fertility or ordering.
Moreover, AligNART (GZ) significantly outperforms previous models for explicit modality reduction except for SNAT on En$\rightarrow$Ro.
In implicit dependency modeling, AligNART (GZ) outperforms Imputer and shows performance comparable to that of the state-of-the-art CTC-based model on En$\leftrightarrow$De by simply augmenting Aligner module to deep-shallow NAT.
In this study, we focus on introducing complete information in word alignments; we do not modify the objective function, which can be explored in the future work.

Table \ref{table:re-scoring_results} shows the BLEU scores with re-scoring decoding strategies of the non-iterative NART models. 
We set $m'=l'=4$, $a=4$, and $b=2$ for 8 candidates.
AligNART outperforms the baselines on En$\rightarrow$De and Ro$\rightarrow$En, and shows performance similar to that of GLAT on De$\rightarrow$En. 
In non-iterative NART for explicit modality reduction, AligNART shows the best performance on En$\leftrightarrow$De and Ro$\rightarrow$En.

\subsection{Analysis of Aligner Components}\label{subsec:analysis_on_aligner_components}
In this section, we investigate the accuracy, example, and ablation results of Aligner components as shown in Table \ref{table:accuracy_on_aligner_components}, \ref{table:qualitative_study}, and \ref{table:alignment_estimation}, respectively. 
Note that we partially provide the ground truth D or P matrices during the accuracy measurement. 

\textbf{Knowledge Distillation} In Table \ref{table:accuracy_on_aligner_components}, a comparison of accuracy between raw and distilled datasets shows that KD significantly decreases \textit{multi-modality} of each component.
After KD, AligNART shows marginally reduced accuracy on the raw dataset, but high prediction accuracy in each component on the distillation set, resulting in increased BLEU scores.

\textbf{Alignment Tool} Before KD, AligNART using \textit{fast align} and GIZA++ have accuracy bottlenecks in permutation and duplication predictors, respectively, as shown in Table \ref{table:accuracy_on_aligner_components}.
The results imply that the alignment tools have different degrees of \textit{multi-modality} on the D, P, and G matrices, which can be explored in the future work.

\textbf{Qualitative Study}
Table \ref{table:qualitative_study} shows an example of addressing the \textit{multi-modality problem}.
Deep-shallow NAT monotonically copies the encoder outputs and suffers from repetition and omission problems.
AligNART (FA) does not show the inconsistency problems thanks to the well-aligned decoder inputs, which significantly reduces the modality of the target distribution. 
We also conducted a case study on predicted alignments and their translations during re-scoring as shown in the Appendix.

\textbf{Ablation Study}
We conduct an analysis of alignment estimation by ablating one of the predictors during inference.
We ablate each module in Aligner by replacing the predicted matrix with an identical matrix $I$.
The results in Table \ref{table:alignment_estimation} indicate that each module in Aligner properly estimates the decomposed information in word alignments.
However, there is an exception in GIZA++ where many-to-one mapping does not exist, resulting in performance equal to that without the grouping predictor.
We observe that AligNART achieves BLEU scores comparable to those of CTC-based models on En$\leftrightarrow$De even with the ground truth word alignments of partial information.

\begin{table}
\centering
\begin{tabular}{l|cc}
\hline
\toprule
\multicolumn{1}{c}{ } &
\multicolumn{2}{c}{WMT14 En-De} \\
 & En$\rightarrow$ & De$\rightarrow$ \\
\hline
FlowSeq (w/o KD) & 18.6 & 23.4 \\
AXE (w/o KD) & 20.4 & \textbf{24.9} \\
Imputer (CTC, w/o KD) & 15.6 & - \\
CTC (w/o KD) & 18.2 & - \\
\hline
NAT (12-1, w/o KD) & 8.5 & 13.3 \\
NAT (12-1, w/ KD) & 18.9 & 23.4 \\
AligNART (FA, w/o KD) & \textbf{20.7} & 24.0 \\
AligNART (GZ, w/o KD) & 18.3 & 23.2 \\
\hline
\end{tabular}
\caption{BLEU scores of non-iterative NART models on WMT14 En-De test set, with or without KD.}
\label{table:KD_studies}
\end{table}

\subsection{Analysis of Modality Reduction Effects}\label{subsec:analysis_on_modality_reduction_effects}

To evaluate the modality reduction effects of AligNART, we conducted experiments on two aspects: BLEU score and token repetition ratio. 
Table \ref{table:KD_studies} shows the BLEU scores on WMT14 En-De. 
For En$\rightarrow$De, AligNART using \textit{fast align} without KD achieves higher BLEU scores than previous models without KD and deep-shallow NAT with KD.
The results indicate that our method is effective even without KD, which is known to decrease data complexity (\citealp{Zhou-2020-Understanding}).
On the other hand, alignments from GIZA++ without KD are more complex for AligNART to learn, resulting in lower BLEU scores than deep-shallow NAT with KD.

\citet{ghazvininejad2020aligned} measured the token repetition ratio as a proxy for measuring \textit{multi-modality}.
The token repetition ratio represents the degree of the inconsistency problem.
In Table \ref{table:token_repetition_studies}, the token repetition ratio of AligNART is less than that of the CMLM-base (\citealp{ghazvininejad-etal-2019-mask}) of 5 iterations, AXE, and GLAT.
We also observe that the decline in the token repetition ratio from Aligner is significantly larger than that from KD.
Combined with the results from Table \ref{table:KD_studies}, alignment information adequately alleviates the token repetition issue even in the case where the BLEU score is lower than that of deep-shallow NAT with KD.

\begin{table}
\centering
\begin{tabular}{l|cc}
\hline
\toprule
\multicolumn{1}{c}{ } &
\multicolumn{2}{c}{WMT14 En-De} \\
 & En$\rightarrow$ & De$\rightarrow$ \\
\hline
Gold test set & 0.04\% & 0.03\% \\
\hline
CMLM-base (5 iterations) & 0.72\% & - \\
AXE & 1.41\% & 1.03\% \\
Imputer (CTC) & 0.17\% & 0.23\% \\
GLAT & 1.19\% & 1.05\% \\
\hline
NAT (12-1, w/o KD) & 33.94\% & 27.78\% \\
NAT (12-1, w/ KD) & 11.83\% & 9.09\% \\
AligNART (GZ, w/o KD) & 0.76\% & 1.33\% \\
AligNART (GZ, w/ KD) & 0.33\% & 0.33\% \\
\hline
\end{tabular}
\caption{Token repetition ratio of NART models on WMT14 En-De test set.}
\label{table:token_repetition_studies}
\end{table}

\subsection{Ablation Study}\label{subsec:ablation_study}
We conduct several extensive experiments to analyze our method further as shown in Table \ref{table:ablation_studies} and \ref{table:alignment_score_filtering}. 
Each of our method consistently improves the performance of AligNART.

\begin{table}
\centering
\begin{tabular}{l|cc}
\hline
\toprule
\multicolumn{1}{c}{ } &
\multicolumn{2}{c}{WMT14 En-De} \\
 & En$\rightarrow$ & De$\rightarrow$ \\
\hline
NAT (12-1) & 18.9 & 23.4 \\
- Cross attention & 17.2 & 21.9 \\
\hline
AligNART (GZ) & 26.4 & 30.4 \\
- Score filtering & 26.2 & 30.0 \\
\hspace{3mm}- Cross attention & 26.1 & 29.9 \\
\hspace{3mm}- 12-1 architecture & 24.9 & 29.1 \\
\hline
\end{tabular}
\caption{Ablation results of deep-shallow NAT and AligNART (GZ) on WMT14 En-De test set.}
\label{table:ablation_studies}
\end{table}

\textbf{Cross Attention} As shown in Table \ref{table:ablation_studies}, we ablate the cross attention module in the decoder to observe the relationship between aligned decoder inputs and alignment learning of the cross attention module. 
We train AligNART and deep-shallow NAT without a cross attention module for comparison. 
AligNART without the cross attention module has a smaller impact on the BLEU score than the deep-shallow NAT. 
The cross attention module is known to learn alignments between source and target tokens (\citealp{bahdanau2015neural}), and the result implies that aligned decoder inputs significantly offload the role of the cross attention module.

\textbf{Deep-shallow Architecture} 
Deep-shallow architecture heavily affects the BLEU scores of AligNART as shown in Table \ref{table:ablation_studies}.
The results indicate that the deep encoder assists alignment estimation, whereas the shallow decoder with aligned inputs has a lower impact on performance degeneration.

\begin{table}
\centering
\begin{tabular}{l|ccccc}
\hline
\toprule
 & 0\% & 1\% & 5\% & 10\% & 20\% \\
\hline
En$\rightarrow$ & 26.2 & 26.1 & 26.4 & 26.2 & 26.2\\
De$\rightarrow$ & 30.0 & 30.2 & 30.4 & 30.4 & 30.1\\
\hline
\end{tabular}
\caption{Alignment score filtering ratio and BLEU scores on WMT14 En-De test set.}
\label{table:alignment_score_filtering}
\end{table}

\textbf{Alignment Score Filtering}
We investigate the trade-off between the alignment score filtering ratio and BLEU score using AligNART (GZ) presented in Table \ref{table:alignment_score_filtering}.
Samples with low alignment scores are more likely to contain noise caused by distilled targets or an alignment tool.
We observe that filtering out of 5\% of the samples improves the BLEU score in both the directions.
Surprisingly, increasing the filtering ratio up to 20\% preserves the performance thanks to the noise filtering capability.

\section{Related Work}\label{sec:relatedwork}
\subsection{Non-iterative NART}\label{subsec:related_work_NART}
After \citet{gu-2018-nonautoregressive} proposed NAT, non-iterative NART has been investigated in various directions to maximize translation speed while maintaining translation quality. 
\citet{shao-etal-2019-retrieving}, \citet{shao-etal-2020-minimizing}, and \citet{ghazvininejad2020aligned} address the limitations of conventional cross entropy based objectives that overly penalize consistent predictions. 
\citet{lee-etal-2018-deterministic}, \citet{ma-etal-2019-flowseq}, \citet{shu2020latent}, and \citet{lee-etal-2020-iterative} introduce latent variables to model the complex dependencies between target tokens.
\citet{saharia-etal-2020-non} and \citet{gu-kong-2021-fully} apply CTC loss to the NMT domain.
\citet{qian-etal-2021-glancing} provide target tokens to the decoder during training using the glancing sampling technique.

\subsection{Alignment in Parallel Generative Models}\label{subsec:related_work_alignment_in_parallel_generative_models}
In other domains, such as text-to-speech (\citealp{ren-2019-fastspeech}; \citealp{kim2020glow}; \citealp{donahue2020end}), a common assumption is a monotonicity in the alignments between text and speech.
Given this assumption, only a duration predictor is required to alleviate the length-mismatch problem between text and speech.
On the other hand, modeling the alignment in the NMT domain is challenging since the alignment contains additional ordering and grouping information.
Our method estimates an arbitrary alignment matrix using \textit{alignment decomposition}.

\subsection{Improving NMT with Enhanced Information}\label{subsec:related_work_improving_NART_with_enhanced_decoder_input}
To alleviate the \textit{multi-modality problem} of NART models, \citet{gu-2018-nonautoregressive}, \citet{akoury-etal-2019-syntactically}, \citet{zhou2020non}, \citet{Ran_2021_guiding}, and \citet{liu-etal-2021-enriching} provide additional sentence information to the decoder.

Alignment is considered as a major factor in machine translation (\citealp{li-etal-2007-probabilistic}; \citealp{zhang-etal-2017-incorporating}).
\citet{alkhouli-etal-2018-alignment} decompose the ART model into alignment and lexical models.
\citet{song-2020-alignment} use the predicted alignment in ART models to constrain vocabulary candidates during decoding. 
However, the alignment estimation in NART is much challenging since the information of decoding outputs is limited.
In NART, \citet{gu-2018-nonautoregressive}, \citet{zhou2020non}, and \citet{Ran_2021_guiding} exploit partial information from the ground truth alignments.
In contrast, we propose the \textit{alignment decomposition} method for effective alignment estimation in NART where we leverage the complete alignment information.
\section{Conclusion and Future Work}\label{sec:conclusion}
In this study, we leverage full alignment information to directly reduce the degree of the \textit{multi-modality} in non-iterative NART and propose an \textit{alignment decomposition} method for alignment estimation.
AligNART with GIZA++ shows performance comparable to that of the recent CTC-based implicit dependency modeling approach on WMT14 En-De and modality reduction capability. 
However, we observe that AligNART depends on the quality of the ground truth word alignments, which can be studied in the future work.
Furthermore, we can study on the combination of AligNART and implicit dependency modeling methods.
\section*{Acknowledgement}\label{sec:acknowledgement}

This work was supported by the National Research Foundation of Korea (NRF) grant funded by the Korea government (Ministry of Science and ICT) [2018R1A2B3001628], the BK21 FOUR program of the Education and Research Program for Future ICT Pioneers, Seoul National University in 2021, AIRS Company in Hyundai \& Kia Motor Company through HKMC-SNU AI Consortium Fund, and Kakao Enterprise.

\bibliography{anthology,custom}

\begin{thebibliography}{39}
\expandafter\ifx\csname natexlab\endcsname\relax\def\natexlab#1{#1}\fi

\bibitem[{Akoury et~al.(2019)Akoury, Krishna, and
  Iyyer}]{akoury-etal-2019-syntactically}
Nader Akoury, Kalpesh Krishna, and Mohit Iyyer. 2019.
\newblock \href {https://doi.org/10.18653/v1/P19-1122} {Syntactically
  supervised transformers for faster neural machine translation}.
\newblock In \emph{Proceedings of the 57th Annual Meeting of the Association
  for Computational Linguistics}, pages 1269--1281, Florence, Italy.
  Association for Computational Linguistics.

\bibitem[{Alkhouli et~al.(2018)Alkhouli, Bretschner, and
  Ney}]{alkhouli-etal-2018-alignment}
Tamer Alkhouli, Gabriel Bretschner, and Hermann Ney. 2018.
\newblock \href {https://doi.org/10.18653/v1/W18-6318} {On the alignment
  problem in multi-head attention-based neural machine translation}.
\newblock In \emph{Proceedings of the Third Conference on Machine Translation:
  Research Papers}, pages 177--185, Brussels, Belgium. Association for
  Computational Linguistics.

\bibitem[{Bahdanau et~al.(2015)Bahdanau, Cho, and Bengio}]{bahdanau2015neural}
Dzmitry Bahdanau, Kyunghyun Cho, and Yoshua Bengio. 2015.
\newblock \href {http://arxiv.org/abs/1409.0473} {Neural machine translation by
  jointly learning to align and translate}.
\newblock In \emph{3rd International Conference on Learning Representations,
  {ICLR} 2015, San Diego, CA, USA, May 7-9, 2015, Conference Track
  Proceedings}.

\bibitem[{Brown et~al.(1993)Brown, Della~Pietra, Della~Pietra, and
  Mercer}]{brown1993mathematics}
Peter~F. Brown, Stephen~A. Della~Pietra, Vincent~J. Della~Pietra, and Robert~L.
  Mercer. 1993.
\newblock \href {https://aclanthology.org/J93-2003} {The mathematics of
  statistical machine translation: Parameter estimation}.
\newblock \emph{Computational Linguistics}, 19(2):263--311.

\bibitem[{Donahue et~al.(2020)Donahue, Dieleman, Binkowski, Elsen, and
  Simonyan}]{donahue2020end}
Jeff Donahue, Sander Dieleman, Mikolaj Binkowski, Erich Elsen, and Karen
  Simonyan. 2020.
\newblock End-to-end adversarial text-to-speech.
\newblock In \emph{International Conference on Learning Representations}.

\bibitem[{Dyer et~al.(2013)Dyer, Chahuneau, and Smith}]{dyer-etal-2013-a}
Chris Dyer, Victor Chahuneau, and Noah~A. Smith. 2013.
\newblock \href {https://aclanthology.org/N13-1073} {A simple, fast, and
  effective reparameterization of {IBM} model 2}.
\newblock In \emph{Proceedings of the 2013 Conference of the North {A}merican
  Chapter of the Association for Computational Linguistics: Human Language
  Technologies}, pages 644--648, Atlanta, Georgia. Association for
  Computational Linguistics.

\bibitem[{Ghazvininejad et~al.(2020)Ghazvininejad, Karpukhin, Zettlemoyer, and
  Levy}]{ghazvininejad2020aligned}
Marjan Ghazvininejad, Vladimir Karpukhin, Luke Zettlemoyer, and Omer Levy.
  2020.
\newblock \href {http://proceedings.mlr.press/v119/ghazvininejad20a.html}
  {Aligned cross entropy for non-autoregressive machine translation}.
\newblock In \emph{Proceedings of the 37th International Conference on Machine
  Learning, {ICML} 2020, 13-18 July 2020, Virtual Event}, volume 119 of
  \emph{Proceedings of Machine Learning Research}, pages 3515--3523. {PMLR}.

\bibitem[{Ghazvininejad et~al.(2019)Ghazvininejad, Levy, Liu, and
  Zettlemoyer}]{ghazvininejad-etal-2019-mask}
Marjan Ghazvininejad, Omer Levy, Yinhan Liu, and Luke Zettlemoyer. 2019.
\newblock \href {https://doi.org/10.18653/v1/D19-1633} {Mask-predict: Parallel
  decoding of conditional masked language models}.
\newblock In \emph{Proceedings of the 2019 Conference on Empirical Methods in
  Natural Language Processing and the 9th International Joint Conference on
  Natural Language Processing (EMNLP-IJCNLP)}, pages 6112--6121, Hong Kong,
  China. Association for Computational Linguistics.

\bibitem[{Graves et~al.(2006)Graves, Fern{\'{a}}ndez, Gomez, and
  Schmidhuber}]{graves-2006-connectionist}
Alex Graves, Santiago Fern{\'{a}}ndez, Faustino~J. Gomez, and J{\"{u}}rgen
  Schmidhuber. 2006.
\newblock \href {https://doi.org/10.1145/1143844.1143891} {Connectionist
  temporal classification: labelling unsegmented sequence data with recurrent
  neural networks}.
\newblock In \emph{Machine Learning, Proceedings of the Twenty-Third
  International Conference {(ICML} 2006), Pittsburgh, Pennsylvania, USA, June
  25-29, 2006}, volume 148 of \emph{{ACM} International Conference Proceeding
  Series}, pages 369--376. {ACM}.

\bibitem[{Gu et~al.(2018)Gu, Bradbury, Xiong, Li, and
  Socher}]{gu-2018-nonautoregressive}
Jiatao Gu, James Bradbury, Caiming Xiong, Victor O.~K. Li, and Richard Socher.
  2018.
\newblock \href {https://openreview.net/forum?id=B1l8BtlCb} {Non-autoregressive
  neural machine translation}.
\newblock In \emph{6th International Conference on Learning Representations,
  {ICLR} 2018, Vancouver, BC, Canada, April 30 - May 3, 2018, Conference Track
  Proceedings}. OpenReview.net.

\bibitem[{Gu and Kong(2021)}]{gu-kong-2021-fully}
Jiatao Gu and Xiang Kong. 2021.
\newblock \href {https://doi.org/10.18653/v1/2021.findings-acl.11} {Fully
  non-autoregressive neural machine translation: Tricks of the trade}.
\newblock In \emph{Findings of the Association for Computational Linguistics:
  ACL-IJCNLP 2021}, pages 120--133, Online. Association for Computational
  Linguistics.

\bibitem[{Guo et~al.(2020)Guo, Xu, and Chen}]{guo-etal-2020-jointly}
Junliang Guo, Linli Xu, and Enhong Chen. 2020.
\newblock \href {https://doi.org/10.18653/v1/2020.acl-main.36} {Jointly masked
  sequence-to-sequence model for non-autoregressive neural machine
  translation}.
\newblock In \emph{Proceedings of the 58th Annual Meeting of the Association
  for Computational Linguistics}, pages 376--385, Online. Association for
  Computational Linguistics.

\bibitem[{Jones et~al.(2001)Jones, Oliphant, Peterson et~al.}]{scipy}
Eric Jones, Travis Oliphant, Pearu Peterson, et~al. 2001.
\newblock \href {http://www.scipy.org/} {{SciPy}: Open source scientific tools
  for {Python}}.

\bibitem[{Kasai et~al.(2020{\natexlab{a}})Kasai, Cross, Ghazvininejad, and
  Gu}]{kasai2020non}
Jungo Kasai, James Cross, Marjan Ghazvininejad, and Jiatao Gu.
  2020{\natexlab{a}}.
\newblock \href {http://proceedings.mlr.press/v119/kasai20a.html}
  {Non-autoregressive machine translation with disentangled context
  transformer}.
\newblock In \emph{Proceedings of the 37th International Conference on Machine
  Learning, {ICML} 2020, 13-18 July 2020, Virtual Event}, volume 119 of
  \emph{Proceedings of Machine Learning Research}, pages 5144--5155. {PMLR}.

\bibitem[{Kasai et~al.(2020{\natexlab{b}})Kasai, Pappas, Peng, Cross, and
  Smith}]{kasai2020deep}
Jungo Kasai, Nikolaos Pappas, Hao Peng, James Cross, and Noah Smith.
  2020{\natexlab{b}}.
\newblock Deep encoder, shallow decoder: Reevaluating non-autoregressive
  machine translation.
\newblock In \emph{International Conference on Learning Representations}.

\bibitem[{Kim et~al.(2020)Kim, Kim, Kong, and Yoon}]{kim2020glow}
Jaehyeon Kim, Sungwon Kim, Jungil Kong, and Sungroh Yoon. 2020.
\newblock \href
  {https://proceedings.neurips.cc/paper/2020/hash/5c3b99e8f92532e5ad1556e53ceea00c-Abstract.html}
  {Glow-tts: {A} generative flow for text-to-speech via monotonic alignment
  search}.
\newblock In \emph{Advances in Neural Information Processing Systems 33: Annual
  Conference on Neural Information Processing Systems 2020, NeurIPS 2020,
  December 6-12, 2020, virtual}.

\bibitem[{Kim and Rush(2016)}]{kim-rush-2016-sequence}
Yoon Kim and Alexander~M. Rush. 2016.
\newblock \href {https://doi.org/10.18653/v1/D16-1139} {Sequence-level
  knowledge distillation}.
\newblock In \emph{Proceedings of the 2016 Conference on Empirical Methods in
  Natural Language Processing}, pages 1317--1327, Austin, Texas. Association
  for Computational Linguistics.

\bibitem[{Kingma and Ba(2015)}]{kingma2015adam}
Diederik~P. Kingma and Jimmy Ba. 2015.
\newblock \href {http://arxiv.org/abs/1412.6980} {Adam: {A} method for
  stochastic optimization}.
\newblock In \emph{3rd International Conference on Learning Representations,
  {ICLR} 2015, San Diego, CA, USA, May 7-9, 2015, Conference Track
  Proceedings}.

\bibitem[{Lee et~al.(2018)Lee, Mansimov, and Cho}]{lee-etal-2018-deterministic}
Jason Lee, Elman Mansimov, and Kyunghyun Cho. 2018.
\newblock \href {https://doi.org/10.18653/v1/D18-1149} {Deterministic
  non-autoregressive neural sequence modeling by iterative refinement}.
\newblock In \emph{Proceedings of the 2018 Conference on Empirical Methods in
  Natural Language Processing}, pages 1173--1182, Brussels, Belgium.
  Association for Computational Linguistics.

\bibitem[{Lee et~al.(2020)Lee, Shu, and Cho}]{lee-etal-2020-iterative}
Jason Lee, Raphael Shu, and Kyunghyun Cho. 2020.
\newblock \href {https://doi.org/10.18653/v1/2020.emnlp-main.73} {Iterative
  refinement in the continuous space for non-autoregressive neural machine
  translation}.
\newblock In \emph{Proceedings of the 2020 Conference on Empirical Methods in
  Natural Language Processing (EMNLP)}, pages 1006--1015, Online. Association
  for Computational Linguistics.

\bibitem[{Li et~al.(2007)Li, Li, Zhang, Li, Zhou, and
  Guan}]{li-etal-2007-probabilistic}
Chi-Ho Li, Minghui Li, Dongdong Zhang, Mu~Li, Ming Zhou, and Yi~Guan. 2007.
\newblock \href {https://www.aclweb.org/anthology/P07-1091} {A probabilistic
  approach to syntax-based reordering for statistical machine translation}.
\newblock In \emph{Proceedings of the 45th Annual Meeting of the Association of
  Computational Linguistics}, pages 720--727, Prague, Czech Republic.
  Association for Computational Linguistics.

\bibitem[{Liu et~al.(2021)Liu, Wan, Zhang, Zhao, and
  Yu}]{liu-etal-2021-enriching}
Ye~Liu, Yao Wan, Jianguo Zhang, Wenting Zhao, and Philip Yu. 2021.
\newblock \href {https://www.aclweb.org/anthology/2021.eacl-main.105}
  {Enriching non-autoregressive transformer with syntactic and semantic
  structures for neural machine translation}.
\newblock In \emph{Proceedings of the 16th Conference of the European Chapter
  of the Association for Computational Linguistics: Main Volume}, pages
  1235--1244, Online. Association for Computational Linguistics.

\bibitem[{Ma et~al.(2019)Ma, Zhou, Li, Neubig, and Hovy}]{ma-etal-2019-flowseq}
Xuezhe Ma, Chunting Zhou, Xian Li, Graham Neubig, and Eduard Hovy. 2019.
\newblock \href {https://doi.org/10.18653/v1/D19-1437} {{F}low{S}eq:
  Non-autoregressive conditional sequence generation with generative flow}.
\newblock In \emph{Proceedings of the 2019 Conference on Empirical Methods in
  Natural Language Processing and the 9th International Joint Conference on
  Natural Language Processing (EMNLP-IJCNLP)}, pages 4282--4292, Hong Kong,
  China. Association for Computational Linguistics.

\bibitem[{Och and Ney(2003)}]{och-ney-2003-systematic}
Franz~Josef Och and Hermann Ney. 2003.
\newblock \href {https://doi.org/10.1162/089120103321337421} {A systematic
  comparison of various statistical alignment models}.
\newblock \emph{Computational Linguistics}, 29(1):19--51.

\bibitem[{Ott et~al.(2019)Ott, Edunov, Baevski, Fan, Gross, Ng, Grangier, and
  Auli}]{ott-etal-2019-fairseq}
Myle Ott, Sergey Edunov, Alexei Baevski, Angela Fan, Sam Gross, Nathan Ng,
  David Grangier, and Michael Auli. 2019.
\newblock \href {https://doi.org/10.18653/v1/N19-4009} {fairseq: A fast,
  extensible toolkit for sequence modeling}.
\newblock In \emph{Proceedings of the 2019 Conference of the North {A}merican
  Chapter of the Association for Computational Linguistics (Demonstrations)},
  pages 48--53, Minneapolis, Minnesota. Association for Computational
  Linguistics.

\bibitem[{Qian et~al.(2021)Qian, Zhou, Bao, Wang, Qiu, Zhang, Yu, and
  Li}]{qian-etal-2021-glancing}
Lihua Qian, Hao Zhou, Yu~Bao, Mingxuan Wang, Lin Qiu, Weinan Zhang, Yong Yu,
  and Lei Li. 2021.
\newblock \href {https://doi.org/10.18653/v1/2021.acl-long.155} {Glancing
  transformer for non-autoregressive neural machine translation}.
\newblock In \emph{Proceedings of the 59th Annual Meeting of the Association
  for Computational Linguistics and the 11th International Joint Conference on
  Natural Language Processing (Volume 1: Long Papers)}, pages 1993--2003,
  Online. Association for Computational Linguistics.

\bibitem[{Ran et~al.(2021)Ran, Lin, Li, and Zhou}]{Ran_2021_guiding}
Qiu Ran, Yankai Lin, Peng Li, and Jie Zhou. 2021.
\newblock \href {https://ojs.aaai.org/index.php/AAAI/article/view/17618}
  {Guiding non-autoregressive neural machine translation decoding with
  reordering information}.
\newblock \emph{Proceedings of the AAAI Conference on Artificial Intelligence},
  35(15):13727--13735.

\bibitem[{Ren et~al.(2019)Ren, Ruan, Tan, Qin, Zhao, Zhao, and
  Liu}]{ren-2019-fastspeech}
Yi~Ren, Yangjun Ruan, Xu~Tan, Tao Qin, Sheng Zhao, Zhou Zhao, and Tie{-}Yan
  Liu. 2019.
\newblock \href
  {https://proceedings.neurips.cc/paper/2019/hash/f63f65b503e22cb970527f23c9ad7db1-Abstract.html}
  {Fastspeech: Fast, robust and controllable text to speech}.
\newblock In \emph{Advances in Neural Information Processing Systems 32: Annual
  Conference on Neural Information Processing Systems 2019, NeurIPS 2019,
  December 8-14, 2019, Vancouver, BC, Canada}, pages 3165--3174.

\bibitem[{Saharia et~al.(2020)Saharia, Chan, Saxena, and
  Norouzi}]{saharia-etal-2020-non}
Chitwan Saharia, William Chan, Saurabh Saxena, and Mohammad Norouzi. 2020.
\newblock \href {https://doi.org/10.18653/v1/2020.emnlp-main.83}
  {Non-autoregressive machine translation with latent alignments}.
\newblock In \emph{Proceedings of the 2020 Conference on Empirical Methods in
  Natural Language Processing (EMNLP)}, pages 1098--1108, Online. Association
  for Computational Linguistics.

\bibitem[{Shao et~al.(2019)Shao, Feng, Zhang, Meng, Chen, and
  Zhou}]{shao-etal-2019-retrieving}
Chenze Shao, Yang Feng, Jinchao Zhang, Fandong Meng, Xilin Chen, and Jie Zhou.
  2019.
\newblock \href {https://doi.org/10.18653/v1/P19-1288} {Retrieving sequential
  information for non-autoregressive neural machine translation}.
\newblock In \emph{Proceedings of the 57th Annual Meeting of the Association
  for Computational Linguistics}, pages 3013--3024, Florence, Italy.
  Association for Computational Linguistics.

\bibitem[{Shao et~al.(2020)Shao, Zhang, Feng, Meng, and
  Zhou}]{shao-etal-2020-minimizing}
Chenze Shao, Jinchao Zhang, Yang Feng, Fandong Meng, and Jie Zhou. 2020.
\newblock \href {https://aaai.org/ojs/index.php/AAAI/article/view/5351}
  {Minimizing the bag-of-ngrams difference for non-autoregressive neural
  machine translation}.
\newblock In \emph{The Thirty-Fourth {AAAI} Conference on Artificial
  Intelligence, {AAAI} 2020, The Thirty-Second Innovative Applications of
  Artificial Intelligence Conference, {IAAI} 2020, The Tenth {AAAI} Symposium
  on Educational Advances in Artificial Intelligence, {EAAI} 2020, New York,
  NY, USA, February 7-12, 2020}, pages 198--205. {AAAI} Press.

\bibitem[{Shu et~al.(2020)Shu, Lee, Nakayama, and Cho}]{shu2020latent}
Raphael Shu, Jason Lee, Hideki Nakayama, and Kyunghyun Cho. 2020.
\newblock Latent-variable non-autoregressive neural machine translation with
  deterministic inference using a delta posterior.
\newblock In \emph{Proceedings of the AAAI Conference on Artificial
  Intelligence}, volume~34, pages 8846--8853.

\bibitem[{Song et~al.(2020)Song, Wang, Yu, Zhang, Huang, Luo, Duan, and
  Zhang}]{song-2020-alignment}
Kai Song, Kun Wang, Heng Yu, Yue Zhang, Zhongqiang Huang, Weihua Luo, Xiangyu
  Duan, and Min Zhang. 2020.
\newblock Alignment-enhanced transformer for constraining nmt with
  pre-specified translations.
\newblock In \emph{AAAI}, pages 8886--8893.

\bibitem[{Sun et~al.(2019)Sun, Li, Wang, He, Lin, and Deng}]{zhiqing2019CRF}
Zhiqing Sun, Zhuohan Li, Haoqing Wang, Di~He, Zi~Lin, and Zhi{-}Hong Deng.
  2019.
\newblock \href
  {https://proceedings.neurips.cc/paper/2019/hash/74563ba21a90da13dacf2a73e3ddefa7-Abstract.html}
  {Fast structured decoding for sequence models}.
\newblock In \emph{Advances in Neural Information Processing Systems 32: Annual
  Conference on Neural Information Processing Systems 2019, NeurIPS 2019,
  December 8-14, 2019, Vancouver, BC, Canada}, pages 3011--3020.

\bibitem[{Sun and Yang(2020)}]{sun2020approach}
Zhiqing Sun and Yiming Yang. 2020.
\newblock \href {http://proceedings.mlr.press/v119/sun20c.html} {An {EM}
  approach to non-autoregressive conditional sequence generation}.
\newblock In \emph{Proceedings of the 37th International Conference on Machine
  Learning, {ICML} 2020, 13-18 July 2020, Virtual Event}, volume 119 of
  \emph{Proceedings of Machine Learning Research}, pages 9249--9258. {PMLR}.

\bibitem[{Vaswani et~al.(2017)Vaswani, Shazeer, Parmar, Uszkoreit, Jones,
  Gomez, Kaiser, and Polosukhin}]{vaswani-2017-attention}
Ashish Vaswani, Noam Shazeer, Niki Parmar, Jakob Uszkoreit, Llion Jones,
  Aidan~N. Gomez, Lukasz Kaiser, and Illia Polosukhin. 2017.
\newblock \href
  {https://proceedings.neurips.cc/paper/2017/hash/3f5ee243547dee91fbd053c1c4a845aa-Abstract.html}
  {Attention is all you need}.
\newblock In \emph{Advances in Neural Information Processing Systems 30: Annual
  Conference on Neural Information Processing Systems 2017, December 4-9, 2017,
  Long Beach, CA, {USA}}, pages 5998--6008.

\bibitem[{Zhang et~al.(2017)Zhang, Wang, Liu, and
  Zhou}]{zhang-etal-2017-incorporating}
Jinchao Zhang, Mingxuan Wang, Qun Liu, and Jie Zhou. 2017.
\newblock \href {https://doi.org/10.18653/v1/P17-1140} {Incorporating word
  reordering knowledge into attention-based neural machine translation}.
\newblock In \emph{Proceedings of the 55th Annual Meeting of the Association
  for Computational Linguistics (Volume 1: Long Papers)}, pages 1524--1534,
  Vancouver, Canada. Association for Computational Linguistics.

\bibitem[{Zhou et~al.(2020{\natexlab{a}})Zhou, Gu, and
  Neubig}]{Zhou-2020-Understanding}
Chunting Zhou, Jiatao Gu, and Graham Neubig. 2020{\natexlab{a}}.
\newblock \href {https://openreview.net/forum?id=BygFVAEKDH} {Understanding
  knowledge distillation in non-autoregressive machine translation}.
\newblock In \emph{8th International Conference on Learning Representations,
  {ICLR} 2020, Addis Ababa, Ethiopia, April 26-30, 2020}. OpenReview.net.

\bibitem[{Zhou et~al.(2020{\natexlab{b}})Zhou, Zhang, Zhao, and
  Zong}]{zhou2020non}
Long Zhou, Jiajun Zhang, Yang Zhao, and Chengqing Zong. 2020{\natexlab{b}}.
\newblock Non-autoregressive neural machine translation with distortion model.
\newblock In \emph{CCF International Conference on Natural Language Processing
  and Chinese Computing}, pages 403--415. Springer.

\end{thebibliography}
\bibliographystyle{acl_natbib}
\clearpage
\appendix

\section*{Appendix}
\section{Mappings in Alignment}
\label{sec:mappings_in_alignment}
\begin{figure}[t]
\centering  
\includegraphics[width=\columnwidth]{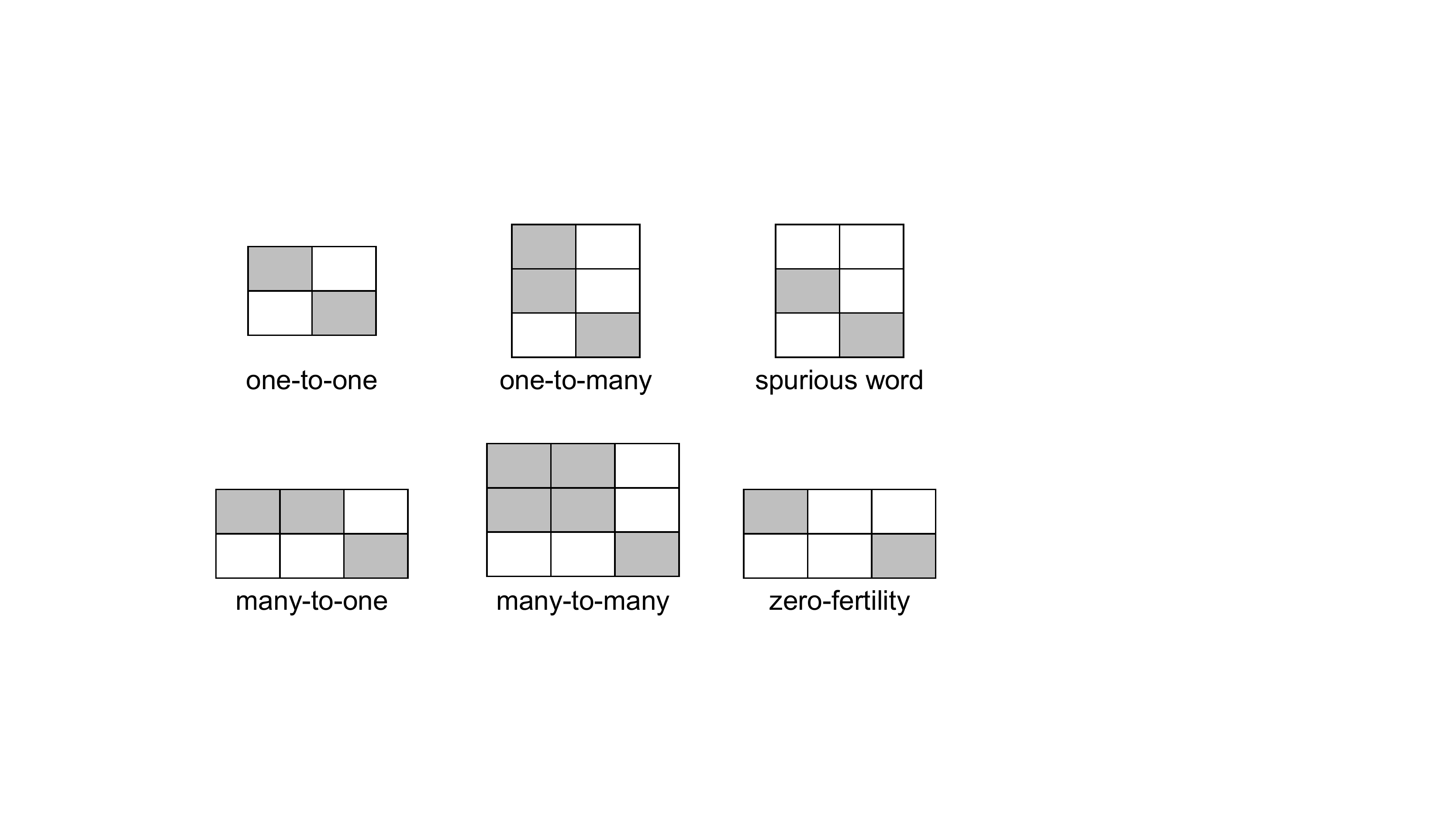}
\caption{Types of mapping in word alignments. Row and colum correspond to the target and source tokens, respectively.}
\label{fig:types_of_mapping_in_word_alignments}
\end{figure}

In general, there are one-to-one, one-to-many, many-to-one, and many-to-many mappings excluding zero-fertility and spurious word cases (see Figure \ref{fig:types_of_mapping_in_word_alignments}).
Distortion and ReorderNAT cannot represent many-to-one, many-to-many, and spurious word cases.
The grouping predictor in AligNART models many-to-one and many-to-many mappings.
The addition of a \textit{spurious} token, which is applied to AligNART (FA), enables us to address the spurious word case, which is explained in Section \ref{subsec:filling_null_rows_in_alignment_matrix}.
During the experiments, we observe that the introduction of a \textit{spurious} token degrades the performance for GIZA++.
We guess the reason of the degradation is that alignment matrix from GIZA++ contains more than two times as many empty rows as that of \textit{fast align} on WMT14 En-De.

\section{Architecture of Aligner}
\label{sec:architecture_of_aligner}
The duplication predictor and grouping predictor modules consist of a convolutional layer, ReLU activation, layer normalization, dropout, and a projection layer, same as the phoneme duration predictor in FastSpeech (\citealp{ren-2019-fastspeech}), which is a parallel text-to-speech model.

The permutation predictor in Aligner consists of three encoder layers: pre-network, query$/$key network, and single-head attention module for the outputs.
Note that the outputs of the pre-network are passed to the query and key networks.
To prevent the predicted permutation matrix from being an identity matrix, we apply a gate function to the last attention module in the permutation predictor to modulate the probabilities of \textit{un-permuted} and \textit{permuted} cases. 
We formulate the output of gated attention as follows:
\begin{equation}
    g=\sigma(Q\cdot u)
\end{equation}
\begin{equation}
    \bar{P}^{pred} = softmax(M + QK^{T})
\end{equation}
\begin{equation}
    P^{pred} = D_g + (I - D_g) \cdot\bar{P}^{pred},
\end{equation}
where $\sigma$ is the sigmoid function and $Q/K$ is the output of the query$/$key network, respectively.
$g$ is the probability of an \textit{un-permuted} case.
$M$ is a diagonal mask matrix, where the values of the diagonal elements are $-inf$. 
$I$ is an identical matrix and $D_g$ is a diagonal matrix with $g$ as the main diagonal.

\begin{figure*}[t]
  \includegraphics[width=\textwidth]{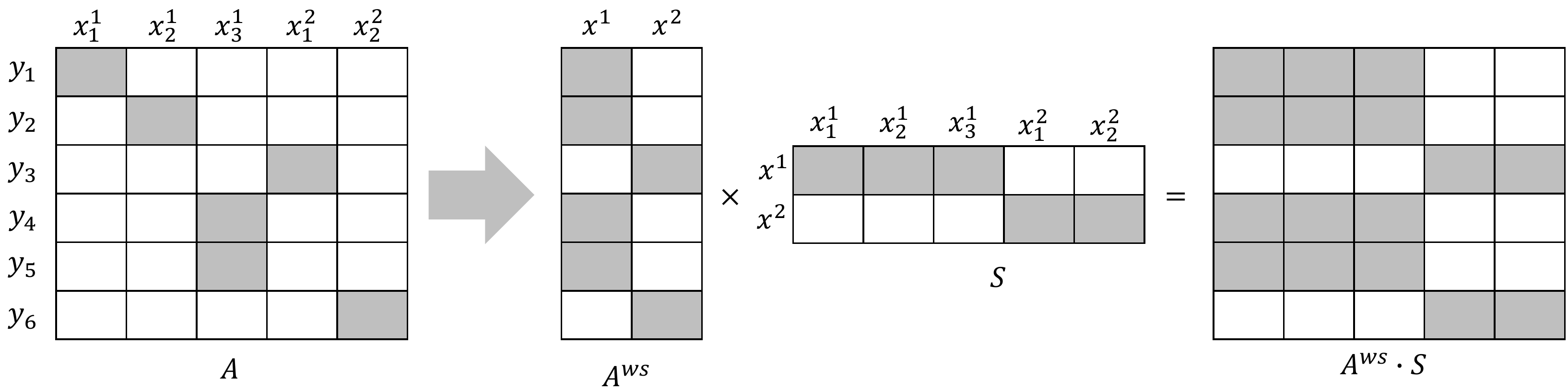}
  \caption{Example of word-to-subword matrix decomposition technique. 
  Row and column correspond to input and output tokens, respectively. 
  $y_i$ denotes the $i$-th subword of the target sentence.
  $x^i$ denotes the $i$-th word of the source sentence and $x^{i}_j$ denotes the $j$-th subword of the $i$-th word of the source sentence.}
  \label{fig:example_of_word_to_subword_matrix_decomposition_technique}
\end{figure*}

\section{Alignment Processing}\label{sec:alignment_processing}
\subsection{Word-to-subword Alignment}\label{subsec:word_to_subword_alignment}
To reduce the complexity of alignment, we further assume that the alignment process is conducted at the word-level.
We decompose the alignment matrix into the source subword to source word matrix $S$ and the source word to target subword matrix $A^{ws}$ as depicted in Figure \ref{fig:example_of_word_to_subword_matrix_decomposition_technique}. 
Since $S$ is always given, $A^{ws}$ is the only target to be learned. First, we derive the source subword to target subword matrix $A$ using the alignment tool. 
$A^{ws}$ is achieved by clipping the maximum value of $A\cdot S^\top$ to 1.
$A^{ws}$ reduces the search space because of the assumption that source tokens duplicate, permute, and group at the word-level.
However, there is a trade-off between the simplicity and resolution of information.
The recovered source subword to target subword matrix $A^{ws}\cdot S$ loses the subword-level information as shown in the rightmost matrix in Figure \ref{fig:example_of_word_to_subword_matrix_decomposition_technique}.

\subsection{Filling Null Rows in Alignment Matrix}\label{subsec:filling_null_rows_in_alignment_matrix} 
The output of the alignment tool usually contains empty rows which means that no aligned source token exists for certain target tokens. 
We select two strategies to fill the null rows: $(i)$ copy the alignment from the previous target token, or $(ii)$ introduce a special \textit{spurious} token. 
For the second strategy, we concatenate a special \textit{spurious} token at the end of the source sentence. 
If the current and previous target tokens belong to the same word, we follow $(i)$. 
The remaining target tokens of the null alignment are aligned to the \textit{spurious} token.

\subsection{Details of Alignment Tool Configuration}\label{subsec:details_on_alignment_tool_configuration}
For \textit{fast align}, we follow the default setting for forward/backward directions and obtain symmetrized alignment with the \textit{grow-diag-final-and} option.
We apply the word-to-subword alignment technique and \textit{spurious} token strategy for null alignments. 
For GIZA++, we apply the word-to-subword alignment technique and copy the alignment from the previous target token for null alignment.
We set the alignment score filtering ratio to 5\%.

\begin{figure*}[!b]
  \includegraphics[width=\textwidth]{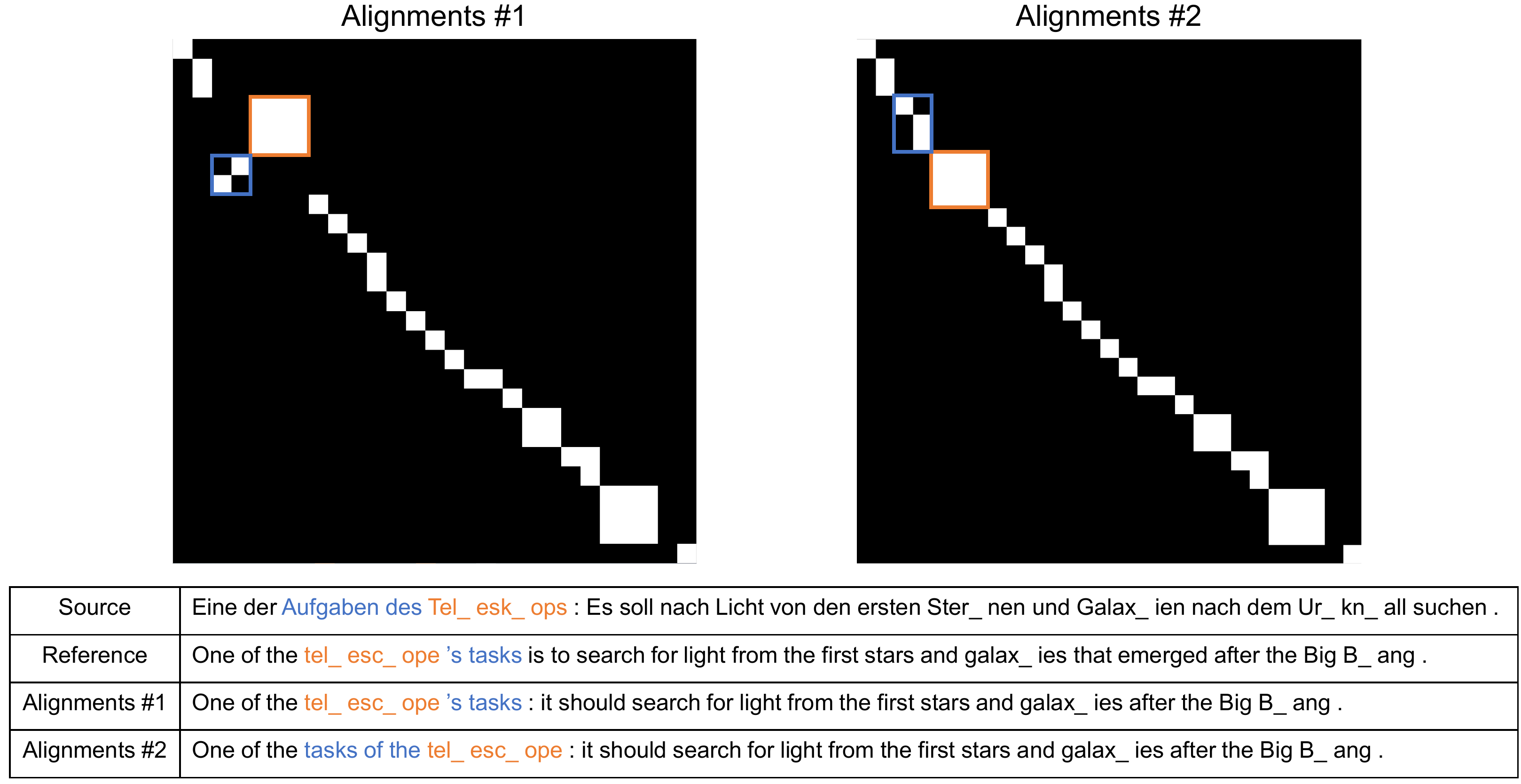}
  \caption{Translation and alignment estimation example on WMT14 De$\rightarrow$En validation set. 
  Tokens matched to the alignment matrix have same colors (blue and orange). 
  The special token "\_" stands for the subword tokenization.}
  \label{fig:case_study}
\end{figure*}

\section{Case Study}\label{sec:case_study}
To analyze various alignments and their translations during re-scoring decoding, we conduct a case study on WMT14 De$\rightarrow$En validation set as shown in Figure \ref{fig:case_study}.
The two translations have different orderings: \textit{the telescope's tasks} and \textit{the tasks of the telescope}.
In this sample, we observe that AligNART $(i)$ can capture non-diagonal alignments, $(ii)$ models multiple alignments, and $(iii)$ translates corresponding to the given alignments.



\end{document}